\documentclass[10pt,twocolumn,letterpaper]{article}

\usepackage{cvpr}              
\usepackage{graphicx}
\usepackage{multirow}
\usepackage{threeparttable}
\usepackage{placeins}

\usepackage[dvipsnames]{xcolor}

\definecolor{cvprblue}{rgb}{0.21,0.49,0.74}
\usepackage[pagebackref,breaklinks,colorlinks,citecolor=cvprblue]{hyperref}
\def\paperID{6} 
\def\confName{CVPR}
\def\confYear{2024}

\title{DTLLM-VLT: Diverse Text Generation for Visual Language Tracking \\ Based on LLM}

\author{
Xuchen Li$^{1}$
Xiaokun Feng$^{1,2}$
Shiyu Hu$^{1,2}$
Meiqi Wu$^{3}$
\\
Dailing Zhang$^{1,2}$
Jing Zhang$^{1}$
Kaiqi Huang$^{1,2,4}$
\\
$^1$CRISE, Institute of Automation, Chinese Academy of Sciences\\
$^2$School of Artificial Intelligence, University of Chinese Academy of Sciences\\
$^3$School of Computer Science and Technology, University of Chinese Academy of Sciences\\
$^4$CAS Center for Excellence in Brain Science and Intelligence Technology\\
\{lixuchen2024, fengxiaokun2022, hushiyu2019, zhangdailing2023, jing\_zhang, kqhuang\}@ia.ac.cn, \\ wumeiqi18@mails.ucas.ac.cn
}

\begin{document}
\maketitle
\begin{abstract}
Visual Language Tracking (VLT) enhances single object tracking (SOT) by integrating natural language descriptions from a video, for the precise tracking of a specified object. By leveraging high-level semantic information, VLT guides object tracking, alleviating the constraints associated with relying on a visual modality. Nevertheless, most VLT benchmarks are annotated in a single granularity and lack a coherent semantic framework to provide scientific guidance. Moreover, coordinating human annotators for high-quality annotations is laborious and time-consuming. To address these challenges, we introduce \textbf{DTLLM-VLT}, which automatically generates extensive and multi-granularity text to enhance environmental diversity. (1) DTLLM-VLT generates scientific and multi-granularity text descriptions using a cohesive prompt framework. Its succinct and highly adaptable design allows seamless integration into various visual tracking benchmarks. (2) We select three prominent benchmarks to deploy our approach: short-term tracking, long-term tracking, and global instance tracking. We offer four granularity combinations for these benchmarks, considering the extent and density of semantic information, thereby showcasing the practicality and versatility of DTLLM-VLT. (3) We conduct comparative experiments on VLT benchmarks with different text granularities, evaluating and analyzing the impact of diverse text on tracking performance. Conclusionally, this work leverages LLM to provide multi-granularity semantic information for VLT task from efficient and diverse perspectives, enabling fine-grained evaluation of multi-modal trackers. In the future, we believe this work can be extended to more datasets to support vision datasets understanding.
\end{abstract}    
\section{Introduction}


\begin{figure}[t]
  \centering
   \includegraphics[width=1\linewidth]{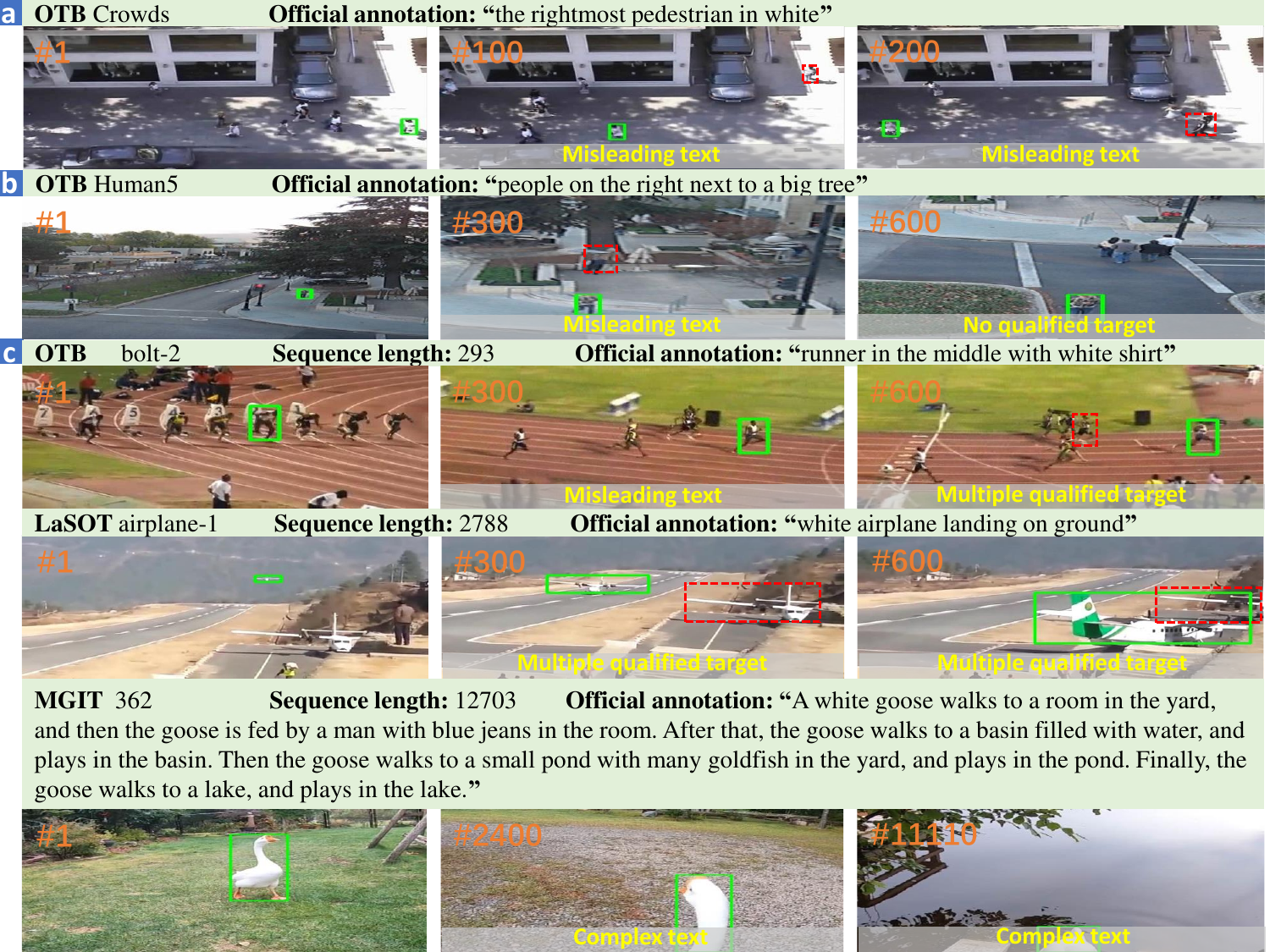}
   \caption{Examples of video content and semantic descriptions on OTB99\_Lang \cite{otb99}, LaSOT \cite{lasot}, and MGIT \cite{mgit} benchmarks. The green bounding box (BBox) indicates ground truth, while the red dashed BBox indicates other objects that satisfy the semantic description. (a) and (b) are short sequences in OTB99\_Lang with simple narrative content. Besides, their semantic annotations mainly describe the first frame, which may misguide the algorithm. (c) Comparison of different text annotations, video length, and content on three benchmarks. The VLT environment is complex, variable and most of them suffer from issues of inconsistent text styles and single annotation granularity.}
   \label{motivation}
\end{figure}

Single object tracking (SOT) is a crucial computer vision task focused on tracking a moving object within a video sequence.
Researchers have consistently observed the limited performance of most trackers in long videos with more complex video content. Moreover, relying solely on a visual modality greatly constrains the versatility of such systems. 
Consequently, several studies have begun providing semantic annotations for the SOT task, leading to the emergence of the visual language tracking (VLT) task. The proposal of VLT task helps the research of SOT to be more human-like and broaden its application prospects. 
Natural language, in contrast to bounding boxes (BBox), provides a more user-friendly and intuitive way of describing objects, allowing for precise descriptions ranging from spatial locations to high-level semantic details to improve tracking performance.
When defining the VLT task, researchers incorporate text annotations from two main viewpoints: 

(1) \textit{Short text annotation.} Representative VLT benchmarks such as OTB99\_Lang \cite{otb99}, TNL2K \cite{tnl2k}, and LaSOT \cite{lasot,lasot21} primarily employ short text. This concise style of description is clear and uncomplicated, facilitating the learning and comprehension of VLT trackers.
The utilization of short text offers the benefit of simplicity and enhanced comprehension for VLT trackers. However, these methods are prone to imprecise semantic descriptions and potential ambiguities. 
As illustrated in Fig.~\ref{motivation} (a) and (b), the description only captures the state of the object at the sequence beginning. As the object moves, the positional constraint in the semantic information becomes misleading.
The reason lies in the benchmark focus primarily on the initial state of the object, neglecting changes in the object's motion throughout the video. Consequently, semantic descriptions may become restrictive later in the sequence. 

(2) \textit{Long text annotation.} MGIT \cite{mgit} adopts a multi-granular semantic annotation strategy from the perspective of more precise semantic descriptions, providing a way to annotate complex spatio-temporal causal relationships in long videos. Compared to other benchmarks, this style exhibits two characteristics: longer text and periodic updates, evolving from simple to dense, detailed descriptions. However, this approach faces challenges like time-intensive text annotations and the need for algorithms with robust text processing and multi-modal alignment capabilities to effectively utilize the information. As shown in Fig.~\ref{motivation} (c), the text in MGIT is overly long and complex.
Clearly, although the motivation of these works is to extend SOT task to multi-modal one to enhance tracking performance, the disparate styles and singular granularity across most studies not only hinder algorithms from achieving the desired outcomes but also escalate the complexity of research on VLT task. 

In summary, diverse motivations in existing research result in varying approaches to integrating textual information. In Fig.~\ref{motivation} (c), the three prominent benchmarks differ in sequence length, text style, and annotation granularity. Imposing a single standard mechanism for VLT research appears impractical, given the inherent flexibility and variability in human comprehension and processing of multi-modal information. 
Humans can adeptly leverage various types of multi-modal information. Rather than enforcing a rigid task format, \textit{\textbf{optimal design should furnish algorithms with comprehensive environmental data to explore their capabilities and limitations}}. 

By offering diverse text descriptions of the environment—encompassing short, long, sparse, and dense formats—and evaluating algorithm performance across these descriptions, we can effectively discern the strengths and weaknesses of existing methods under different semantic granularities, thereby guiding the enhancement of multi-modal algorithms. 
What excites us is that the Large Language Model (LLM) can facilitate the achievement of this goal. By seamlessly integrating the LLM into the text generation process, we can offer a varied multi-modal environment conducive to VLT research.

Our work focuses on the aforementioned motivations and designs DTLLM-VLT to achieve diverse text generation for tracking datasets. Specifically, we combine text length and generation density to form four granularities with a uniform style. Based on this, we select MMTrack \cite{mmtrack}, a state-of-the-art (SOTA) VLT tracker, for experimental analysis to verify the impact of diverse texts on algorithm performance. The experimental results not only demonstrate that this diversified environment can assist in fine-grained evaluation and analysis of algorithm capabilities but also suggest the possibility of further enhancing the multi-modal learning capabilities of algorithms using generated data in the future.

The contributions of this paper can be summarized in the following three aspects:
    
    
    
    \begin{itemize}
    \item 
    We develop DTLLM-VLT, a model based on LLM, aimed at efficiently generating high-quality scientific text for tracking datasets at scale. DTLLM-VLT can seamlessly apply to various tracking tasks.
    
    \item 
    We generate diverse text for three prominent VLT benchmarks, addressing four levels of granularity. This approach overcomes the limitations of previous benchmarks, which focused on a single granularity and lacked a unified semantic framework.
    
    \item 
    We conduct an experimental analysis to evaluate the impact of diverse texts on algorithm performance. The results highlight the benefits of a diversified environment and indicate the potential for enhancing multi-modal learning through generated text data.

    \end{itemize}

\section{Related Work}


\begin{table*}[t!]
    \centering
    \caption{Summary of current popular tracking benchmarks and Comparison number of language description between official and our generated text. \textit{Italics} indicate automatic generation. We provide far more diverse semantic information than the original annotations for representative environments.}
    \label{vldataset}
    \begin{threeparttable}
    \begin{tabular}{c|cc|ccccc}
      \hline
      \multirow{2}{*}{{Dataset}} &\multicolumn{2}{c|}{{Number of Videos}}  &\multicolumn{5}{c}{{Number of Language Description}}  \\
      \cline{2-8}
      &Train &Evaluation & {Official} & \textit{{Dense Concise}} & \textit{{Dense Detailed}} & \textit{{Initial Concise}} & \textit{{Initial Detailed}} \\
      \hline
      OTB99\_Lang \cite{otb99}  & 51 & 48 & 99 & \textit{596} & \textit{596} & \textit{99} & \textit{99}      \\
      LaSOT \cite{lasot}  & 1,120 & 280 & 1,400 & \textit{35.2K} & \textit{35.2K} & \textit{1,400} & \textit{1,400}        \\
      TNL2K \cite{tnl2k}  & 1,300 & 700 & 2,000 & \textit{12.4K} & \textit{12.4K} & \textit{2,000} & \textit{2,000}        \\
      MGIT\tnote{1} \cite{mgit}   & 105 & 45 & 1,753 & \textit{16.1K} & \textit{16.1K} & \textit{120} & \textit{120}          \\
      \hline
    \end{tabular}
    \begin{tablenotes}
        \footnotesize
        \item[1] As the ground truth of the MGIT \cite{mgit} test set is not open-sourced, we only generated text for 120 video of the training and validation sets.
    \end{tablenotes}
    \end{threeparttable}
\end{table*}

\subsection{Single Object Tracking Benchmark}
The SOT task involves initializing and tracking a specific object within a video sequence. It begins by identifying the object through its BBox in the first frame and then proceeds to locate and follow the object across subsequent frames.
Since 2013, several benchmarks such as OTB \cite{otb50,otb100} and VOT \cite{evaluate,vot15} have been introduced, providing standardized datasets and scientific evaluation mechanisms to support SOT research. However, with the advancements in deep learning techniques, these short-term and small-scale benchmarks have faced challenges in adequately accommodating data-driven trackers. Consequently, researchers have started designing larger-scale datasets such as GOT-10k \cite{got10k} and TrackingNet \cite{trackingnet}. Additionally, efforts have been made to gather data featuring long videos, leading to the creation of long-term tracking benchmarks like OxUvA \cite{oxuva} and VOT\_LT \cite{vot18,vot19}. Some work has also focused on SOT in drone scenarios, such as BioDrone \cite{biodrone}, a vision benchmark for SOT based on bionic drones.
Recently, researchers have acknowledged that traditional approaches to both short-term and long-term tracking are based on the premise of constant movement, a factor that restricts testing to situations involving a single camera view and a static scene. To expand beyond these limitations, they have introduced the global instance tracking task along with a novel benchmark called VideoCube \cite{git}, which enables the tracking of arbitrary moving objects in various types of videos.
To scientifically evaluate the performance of trackers under different challenging factors, researchers have introduced SOTVerse \cite{sotverse}, a user-defined space for SOT task.

\subsection{Visual Language Tracking Benchmark}
While visual benchmarks have undergone significant evolution over the past decades, benchmarks integrating visual and semantic information, known as VLT benchmarks, have only recently gained traction. OTB99\_Lang \cite{otb99} stands out as the first VLT benchmark, enhancing sequences from the OTB100 \cite{otb100} benchmark with additional natural language descriptions. However, the limited scale of the dataset has hindered the widespread adoption of the VLT task. Subsequently, the release of LaSOT \cite{lasot,lasot21}, a long-term tracking benchmark with natural language annotations, marked a significant development. Concurrently, researchers introduced the TNL2K \cite{tnl2k} benchmark in the same year, aiming to enhance object tracking flexibility and accuracy through text descriptions. Following these efforts, researchers proposed a new multi-modal benchmark named MGIT \cite{mgit}, which fully represents the complex spatio-temporal and causal relationships present in long narrative content through a multi-granular annotation strategy. These three benchmarks have enriched the pool of available data and facilitated the development of various VLT trackers. 

\subsection{Algorithms for Visual Language Tracking}
VLT emerges as a burgeoning multi-modal task aiming to achieve tracking by leveraging both a language description and an initial template patch. Following the principle of similarity-matching, most existing VLT methods \cite{vlt,transnlt,transvlt,ctrtnl,arxiv19,dat,rttnld} utilize language descriptions and template patches as references to identify the most similar object in the search frame. Among these methods, SNLT \cite{snlt} presents an adaptable language-based region proposal network that improves tracking accuracy by employing a dynamic aggregation mechanism. Meanwhile, MMTrack \cite{mmtrack} introduces a streamlined and effective tracking method, treating the VLT task as a sequence of token generation. However, these methods often fail to capture the dynamic properties of the object, which becomes a critical issue for robust tracking when the object's appearance undergoes significant changes. To overcome this shortcoming, some VLT trackers have begun to integrate temporal data to establish a more dynamic reference. For instance, GTI \cite{gti} and AdaSwitcher \cite{tnl2k} identify object by merging tracking and localization outcomes at every time interval. JointNLT \cite{jointnlt} also takes a step towards this by including temporal information as queries during the prediction phase.


Most benchmarks for VLT provide only one natural language description per video. Additionally, the existing benchmarks suffer from inconsistent text annotation styles, leading to varied mechanisms for incorporating text information. These discrepancies hinder algorithm evaluation and comprehension of video content. Moreover, these works all provide semantic information in the form of manually annotated data, which is a time-consuming and labor-intensive process.

\begin{figure*}[ht!]
  \centering
    \includegraphics[width=1\linewidth]{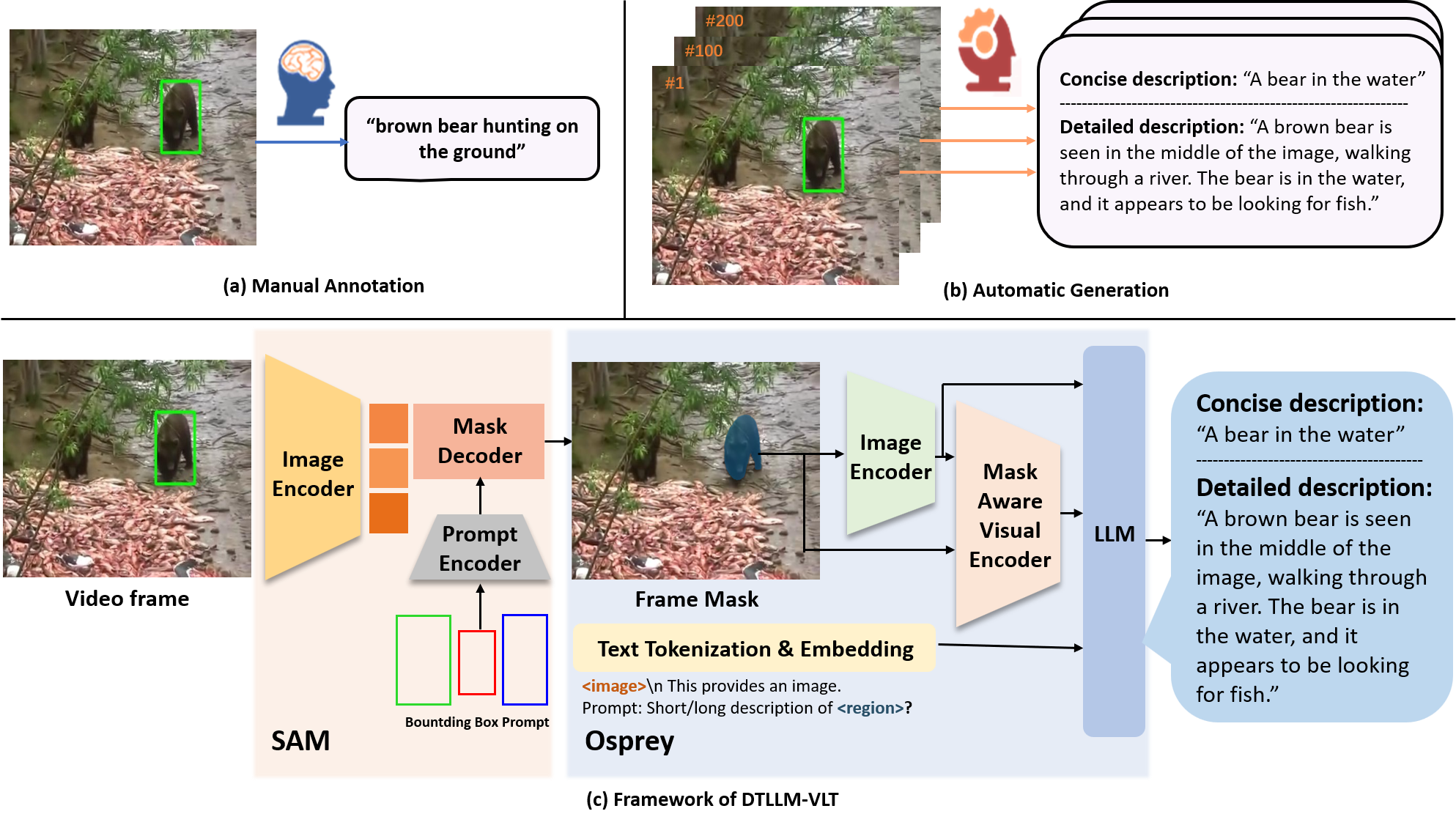}
    \caption{Comparison of Manual Annotation and Automatic Generation and Framework of DTLLM-VLT. (a) Manual annotation relies on human labor, only provides one text annotation for each video segment, and cannot guarantee a uniform style. The cost of large-scale annotation is too high. (b) Automatic Generation can generate diverse text on a large-scale in a unified style. (c) The DTLLM-VLT can provide dense concise/detailed text generation based on given video frames and BBox of object.}
    \label{method}
\end{figure*}

\section{Text Generation by LLM}




To provide diverse text generation for VLT datasets under a unified prompt framework and provide algorithms with more scientific text for evaluation and understanding video content, we implement DTLLM-VLT to offer large-scale automatic diverse generated text.


\begin{figure}[ht!]
  \centering
   \includegraphics[width=1\linewidth]{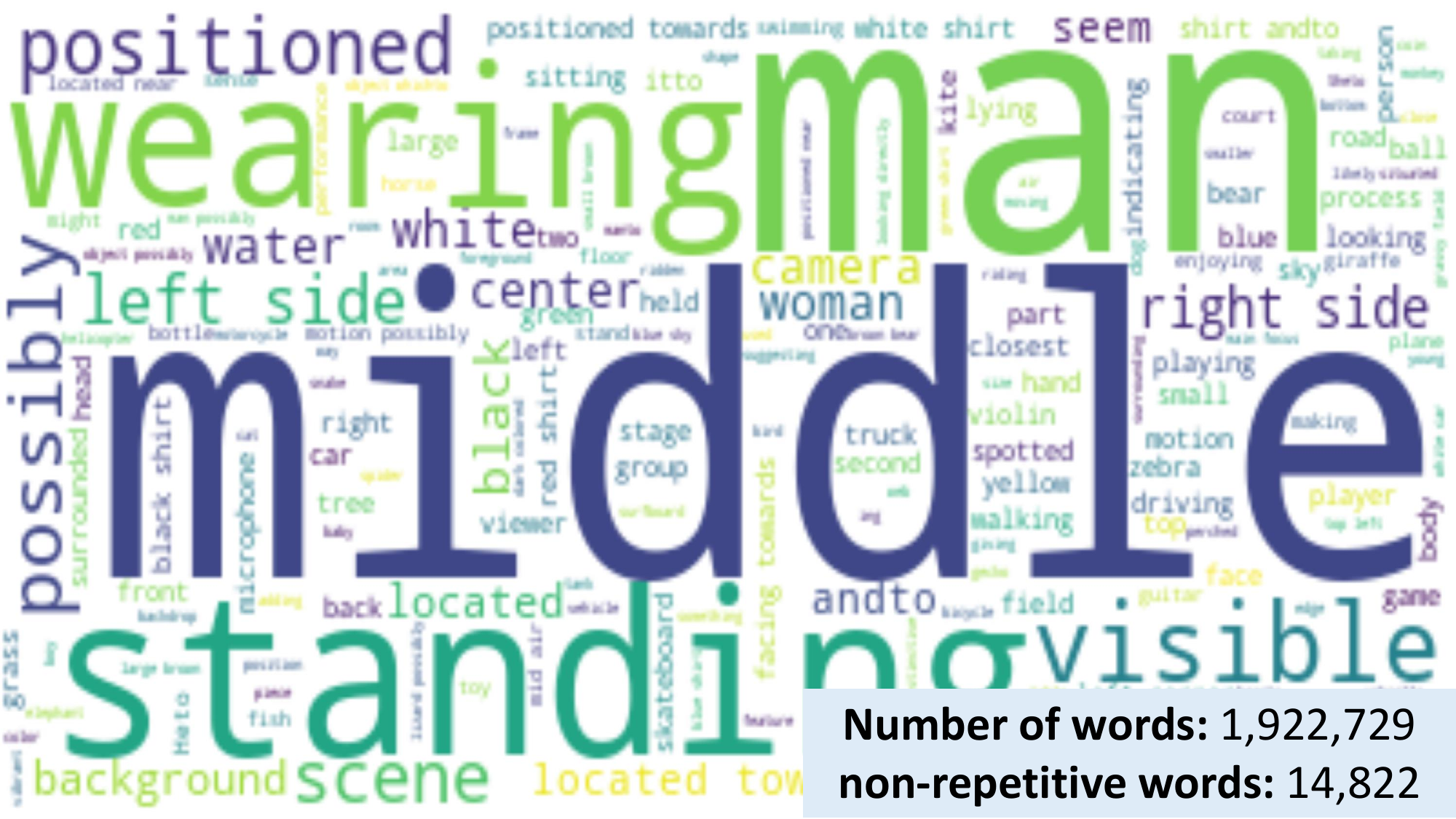}

   \caption{The word cloud of semantic descriptions and word count statistics.}
   \label{word}
\end{figure}

\begin{figure*}[ht!]
  \centering
    \includegraphics[width=1\linewidth]{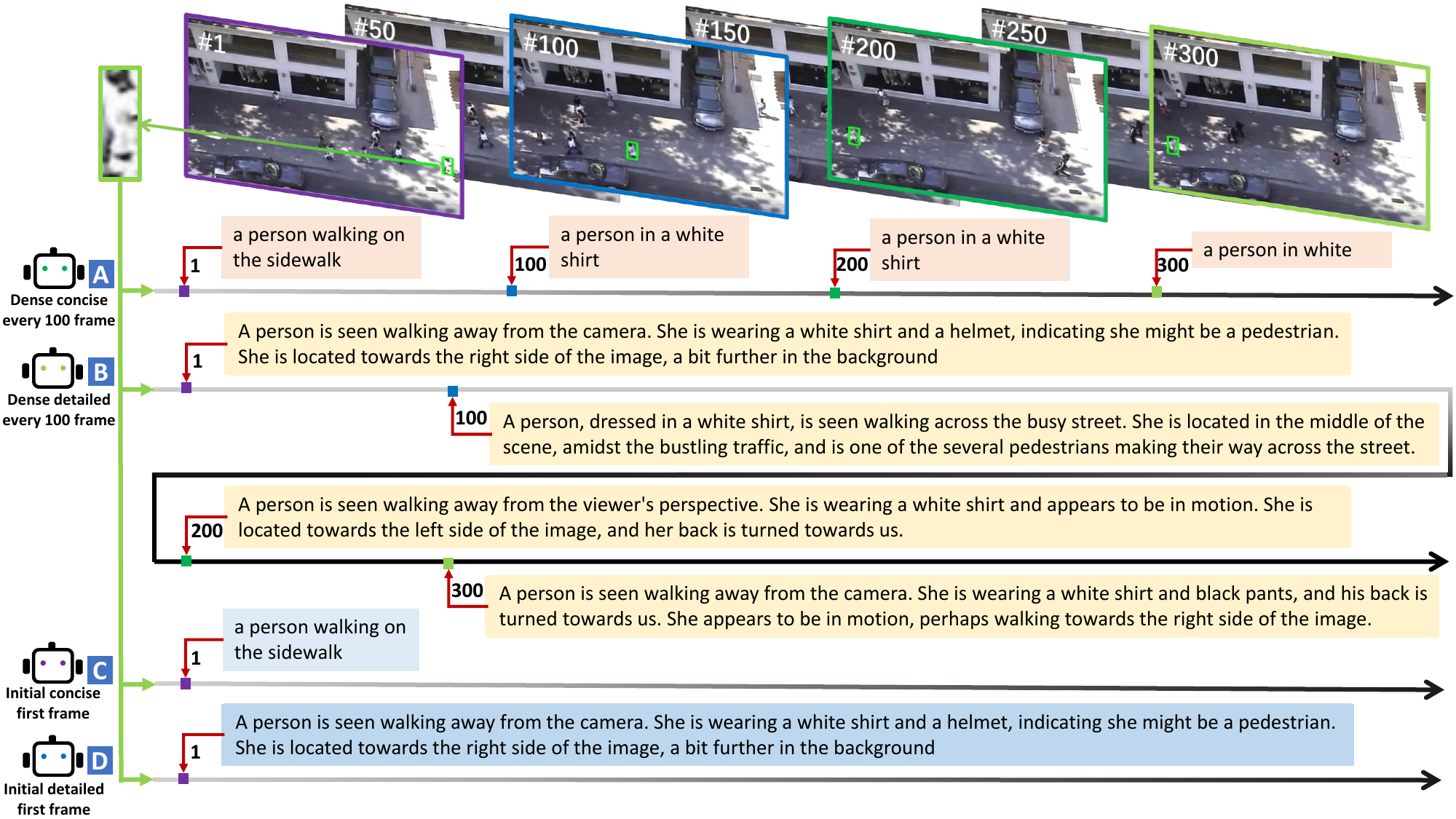}
    \caption{Examples of the four types of generated text. We provide four different natural language descriptions for each video. The object to be tracked is determined in the first frame and does not change throughout the video sequence.}
    \label{annotation}
\end{figure*}

\subsection{Generation Strategy}
The volume and linguistic annotations of the VLT dataset determine the quality and generality of learned visual language representations. 
Table~\ref{vldataset} illustrates that the dataset comprises only 3,649 videos, specifically 1,400 from LaSOT \cite{lasot}, 2,000 from TNL2K \cite{tnl2k}, 99 from OTB99\_Lang \cite{otb99}, and 150 from MGIT \cite{mgit}, which are used for training and testing. These videos are accompanied by 5,252 official text descriptions. However, this amount of data is deemed insufficient for algorithms to effectively learn.

These official annotation suffers from inconsistency in style, and are only able to describe short-term changes for the object. The varying annotation styles of the text descriptions make it difficult for trackers to learn general visual language information, resulting in a significant performance drop when inferring on new videos with non-official annotations or different language description styles. Moreover, inaccurate text descriptions hinder object tracking, turning natural language annotations into a hindrance rather than a support.

To enhance the accuracy and generality, we propose DTLLM-VLT, which generates text in a consistent style for four datasets, establishing a robust foundation for VLT. This generation approach can be expanded to additional VLT datasets and even applied to text generation in SOT datasets.



\textbf{Initial and dense text descriptions.}
Following the text annotations method in OTB99\_Lang \cite{otb99} and TNL2K \cite{tnl2k}, we generate text for the initial frame of each video. Additionally, given that 4 seconds marks the threshold between human instant memory and short-term memory \cite{humanmemory,memory,human}, we consider the worst situation and infer that the algorithm lacks an efficient memory system. Consequently, at 25 FPS, equating to every 100 frames in 4 seconds, we supply the algorithm with relevant generated text. We posit that this update frequency optimally sustains the memory state of algorithm and enhances tracking performance.

\textbf{Concise and detailed text descriptions.}
For the algorithm, if the BBox already sufficiently describes the temporal and spatial changes of the object, the text descriptions should focus on providing essential semantic details like the category and positions of the object. In cases where the BBox lacks sufficient information for effective learning by the tracker, more elaborate texts are necessary to compensate for the missing temporal and spatial relationships. Consequently, we generate two types of text descriptions: concise and detailed. As illustrated in Fig.~\ref{method}, the concise text conveys essential information about the object, such as its category (\textit{bear}) and position (\textit{in the water}), while the detailed text includes additional spatio-temporal details like color, relative position, and actions.


\subsection{DTLLM-VLT}
The traditional VLT datasets rely on manual text annotations, as shown in Fig.~\ref{method} (a), providing a corresponding natural language description for each video. This method incurs high annotation costs, lacks uniformity in style, involves a single annotation granularity, and cannot be used for large-scale data annotation. To address these issues, we design DTLLM-VLT based on SAM \cite{sam} and Osprey \cite{osprey}, which can provide large-scale and diverse text generation like Fig.~\ref{method} (b).

The framework of the DTLLM-VLT is illustrated in Fig.~\ref{method} (c). Input video frames and corresponding object BBox, SAM \cite{sam} utilizes image encoder, prompt encoder, and mask decoder to obtain masks of the corresponding object and then input the video frames and mask into Osprey \cite{osprey}. Osprey encodes the images and masks, combines with preset prompts, and generates concise and detailed descriptions of the corresponding object through LLM \cite{vicuna,llama}. Through this approach, we can generate large-scale, diverse granularities, and uniform style text for SOT and VLT datasets at very low costs.




\subsection{Generation Analysis}
Combining the aforementioned strategies, we offer four granularities of natural language descriptions for each video, namely initial concise description, initial detailed description, dense concise description, and dense detailed description, as illustrated in Fig.~\ref{annotation}. Our goal is to incorporate multiple granularities of text to enrich the environment for algorithm to learn and evaluate, while also providing guidance for algorithm design and model optimization.

Leveraging the DTLLM-VLT, we generate text descriptions comprising 7,238 initial descriptions (3,619 concise and 3,619 detailed descriptions each) and 128.4K dense descriptions (64.3K concise and 64.3K detailed descriptions each). Our dense texts are 24.4 times the quantity of the official annotations. Further details regarding the number of semantic descriptions are presented in Table~\ref{vldataset}. The semantic descriptions contain 1.9M words with 14.8K non-repetitive words. The vocabulary is rich, allowing for a comprehensive description of changes in the object during the tracking process. Word cloud and more detailed analyses have been illustrated in Fig.~\ref{word}.

\subsection{Speed and Memory Usage}
We generate diverse text for visual language tracking datasets on RTX-3090 GPUs, with approximately 16GB of VRAM usage. It takes about 2 seconds to generate a text entry for each frame. 

Compared to manual annotation, DTLLM-VLT can generate texts of various granularities for large-scale tracking datasets in a short period of time. And it can seamlessly apply to various tracking tasks.





\section{Experimental Results}
\begin{table*}[h!]
    \centering
    \caption{Comparison with testing directly on three popular benchmarks: OTB99\_Lang \cite{otb99}, MGIT \cite{mgit}, and LaSOT \cite{lasot}. 
    The best two results are highlighted in {\color{red}red} and {\color{blue}blue}, respectively.}
    \label{results}
    \begin{tabular}{l|ccc|ccc|ccc}
    \toprule
     \multicolumn{1}{c|}{\multirow{2}{*}{Method}}
      & \multicolumn{3}{c|}{OTB99\_Lang \cite{otb99}} & \multicolumn{3}{c|}{MGIT \cite{mgit}} & \multicolumn{3}{c}{LaSOT \cite{lasot}} \\ \cline{2-10}
      & AUC & P${_{\text{Norm}}}$ & P & AUC & P${_{\text{Norm}}}$ & P & AUC & P${_{\text{Norm}}}$ &P  \\
      \midrule
      Official          & 69.0 & 82.0 & 89.5 & 73.5 & 77.2 & 54.3 & {\color{red}69.9} & {\color{red}82.2} & {\color{red}75.7} \\ 
      Initial Concise   & {\color{red}70.6} & {\color{red}84.2} & {\color{red}91.1} & {\color{blue}73.9} & {\color{blue}77.8} & {\color{blue}54.9} & 69.0 & 81.1 & 74.7 \\ 
      Initial Detailed  & 68.0 & 81.5 & 88.4 & 72.7 & 76.2 & 53.4 & 68.7 & 80.7 & 74.4 \\
      Dense Concise     & {\color{blue}70.2} & {\color{blue}84.0} & {\color{blue}90.8} & {\color{red}74.2} & {\color{red}77.9} & {\color{red}55.0} & {\color{blue}69.1} & {\color{blue}81.3} & {\color{blue}74.8} \\
      Dense Detailed    & 68.6 & 82.4 & 89.4 & 72.9 & 76.6 & 53.5 & 69.0 & 81.1 & 74.7 \\ 
    \bottomrule
    \end{tabular} 
\end{table*}

\begin{table*}[h!]
    \centering
    \caption{Comparison with retraining and testing respectively on three popular benchmarks: OTB99\_Lang \cite{otb99}, MGIT \cite{mgit}, and LaSOT \cite{lasot}. 
    The best two results are highlighted in {\color{red}red} and {\color{blue}blue}, respectively.}
    \label{results_retrain}
    \begin{tabular}{l|ccc|ccc|ccc}
    \toprule
     \multicolumn{1}{c|}{\multirow{2}{*}{Method}}
      & \multicolumn{3}{c|}{OTB99\_Lang \cite{otb99}} & \multicolumn{3}{c|}{MGIT \cite{mgit}} & \multicolumn{3}{c}{LaSOT \cite{lasot}} \\ \cline{2-10}
      & AUC & P${_{\text{Norm}}}$ & P & AUC & P${_{\text{Norm}}}$ & P & AUC & P${_{\text{Norm}}}$ &P  \\
      \midrule
      Official         & 69.0 & 82.0 & 89.5 & 73.5 & 77.2 & 54.3 & {\color{red}69.9} & {\color{red}82.2} & {\color{red}75.7} \\ 
      Initial Concise  & 70.0 & 84.3 & 90.5 & 73.6 & 77.4 & 54.2 & 69.6 & 81.8 & 75.4 \\ 
      Initial Detailed & {\color{blue}70.3} & {\color{blue}85.6} & {\color{blue}91.4} & {\color{blue}74.1} & {\color{blue}78.3} & {\color{blue}54.5} & 69.4 & 81.5 & 75.1 \\
      Dense Concise    & {\color{red}71.3} & {\color{red}86.0} & {\color{red}92.5} & 74.0 & 77.6 & 54.2 & 69.5 & 81.6 & 75.3 \\
      Dense Detailed   & 69.8 & 84.8 & 90.6 & {\color{red}74.4} & {\color{red}78.5} & {\color{red}54.6} & {\color{blue}69.8} & {\color{blue}82.1} & {\color{blue}75.6} \\ 
    \bottomrule
    \end{tabular} 
\end{table*}


\subsection{Datasets and Evaluation Methods}
\textbf{Datasets.}
We selected three representative datasets, OTB99\_Lang \cite{otb99}, LaSOT \cite{lasot}, and MGIT \cite{mgit}, for evaluating short-term tracking, long-term tracking, and global instance tracking task. 
OTB99\_Lang \cite{otb99} and LaSOT \cite{lasot} are expanded from the traditional SOT benchmark by adding language annotations. OTB99\_Lang serves as a representative dataset for short-term tracking task, providing a text description for the initial frame of each video sequence. LaSOT is a representative dataset for long-term tracking task. Its text annotations only describe the appearance of the target, omitting relative positions.
MGIT \cite{mgit} is a novel large-scale benchmark specifically tailored for the global instance tracking task. Text annotations of each sequence contain complex spatio-temporal causal relationships with a multi-granular annotation strategy.

\textbf{Evaluation Methods.}
As shown in Fig.~\ref{annotation}, we follow generation granularities to design various mechanisms.
We select a SOTA visual language tracker, MMTrack \cite{mmtrack} as a baseline model and evaluate it on three benchmarks (as shown in Table~\ref{results} and Table~\ref{results_retrain}). Compared with other algorithms, MMTrack \cite{mmtrack} does not impose restrictions on the length of the text and does not truncate excessively long text. Additionally, it unifies the VLT task as a form of token generation, which is more conducive to learning visual language 
information.

To fairly compare the tracking performance on three datasets, we directly use the officially provided weights to test with the official annotations, initial concise texts, initial detailed texts, dense concise texts, and dense detailed texts. We also retrain and test the model under the corresponding settings to evaluate Area Under the Curve (AUC), tracking precision (P), and normalized precision (P${_{\text{Norm}}}$). 

\subsection{Tracking Results}
We evaluate MMTrack \cite{mmtrack} on three benchmarks, including OTB99\_Lang \cite{otb99}, MGIT \cite{mgit}, and LaSOT \cite{lasot} with five text granularities to evaluate the influence of diverse generated text on tracking performance. All experiments employ joint language and BBox initialization.



\subsubsection{Testing Directly}
We directly use the model provided by the official for testing, and the test results are as shown in Table~\ref{results}. 

\textbf{Short-term tracking.} In Table~\ref{results}, when comparing results on OTB99\_Lang \cite{otb99}, which only provides the text description of the initial frame and will interfere with the tracking of the object in the later stage, our initial concise text achieves gains of 1.6 \%, 2.2 \%, and 1.6 \% in area under the curve, normalized precision, and precision score, respectively. At the same time, we find that dense concise text also helps improve tracking performance, for example, our generated text achieves improvements of 1.2 \% in the area under the curve.
We think that the short-term tracking datasets represented by OTB99\_Lang \cite{otb99}, their BBox can effectively describe the temporal and spatial relationships in the visual modality. If only the text from the initial frame is used and cannot describe the temporal and spatial relationships of the object in the following frame, it will cause significant interference. The same problem arises in our detailed initial concise/dense text description testing. In this case, the text only needs to be as concise as possible to assist in improving tracking performance. 

\textbf{Long-term tracking.} The official text annotation of LaSOT \cite{lasot} only describes the appearance of the object, ignoring the relative position. Compared to OTB99\_Lang \cite{otb99}, the text description of the object is more accurate. Compared with MGIT \cite{mgit}, there is no excessive interference from relative position information. It represents a balance between the two and is most in line with the current algorithm learning method. Therefore, the test performance with official annotation is the best. However, we believe that for long-term tracking, providing only a single sentence of text is not conducive to algorithm learning. And the spatial relationships of the object are crucial. When there are large-scale and diverse VLT datasets and better approaches to enhancing video understanding capabilities of algorithm, this situation observed in LaSOT \cite{lasot} will soon change.

\textbf{Global instance tracking.} The same situation as OTB99\_Lang \cite{otb99} appeared on MGIT \cite{mgit}, that is, the performance is improved when tested under initial/dense concise text annotations. Particularly, dense concise annotation excels over the official text, surpassing it by 0.7 \%, 0.7 \%, and 0.7 \% in area under the curve, normalized precision, and precision score, respectively. MGIT \cite{mgit} provides high-quality, multi-granularity long texts containing complex temporal and spatial relationships. From the test results, we think that the handling of long texts and multimodal alignment in the current algorithm requires improvement, as it fails to fully leverage temporal and spatial relationships. Therefore, concise text can actually help improve performance. However, temporal and spatial information are crucial for long-term tracking and global instance tracking. When the temporal-spatial information of the BBox cannot stably determine the object, detailed text is needed to provide additional high-level semantic information to identify the object. 

Through direct testing and comparison of tracking performance under different texts, it has been observed that the variation in texts has a significant impact on tracking performance. The largest performance difference reached 2.2\% in normalized precision on the OTB99\_Lang dataset.



\subsubsection{Retraining and Testing Respectively}
As mentioned earlier, when the dataset text becomes denser and more accurate, it can compensate for BBox shortcomings. The algorithm gains additional knowledge through text updates, potentially improving performance. Therefore, we retrained and tested MMTrack \cite{mmtrack} using varied generated texts, with tracking results shown in Table~\ref{results_retrain}.

\textbf{Short-term tracking.} It can be seen that on the OTB99\_Lang \cite{otb99} benchmark, the testing results after retraining with dense concise text have shown further improvement. Compared with the official text, it gains 2.3 \%, 4.0 \%, and 3.0 \% in area under the curve, normalized precision, and precision score, respectively. This indicates that providing dense concise text on short-term datasets can further improve tracking performance. It also reflects the capability of the current algorithm to achieve better tracking even when provided with more accurate text, without the need for matching learning methods. However, we believe that the current method of training algorithms to memorize high-frequency text for enhancing memory capabilities still needs improvement, the potential of text has not been fully exploited yet.

\textbf{Long-term tracking.} The results on the LaSOT \cite{lasot} benchmark show that official annotations are still more advantageous for tracking. However, after retraining, the results on dense detailed text are only 0.1 \% from the optimal results, indicating an improvement in the algorithm's understanding of dense text compared to direct testing, but it is still unable to fully learn all temporal and spatial information.

\textbf{Global instance tracking.} The test results after retraining based on different texts show that the algorithm can improve its tracking ability on the MGIT \cite{mgit} benchmark by learning from dense detailed text, which differs from the results of direct testing. For global instance tracking task, it is beneficial for tracking if the algorithm can learn more comprehensive temporal and spatial relationships. 

Comparing the above results, we can draw the following insights: 

(1) \textbf{The existing algorithm tends to learn and understand short text.} The results of direct testing show that concise text is more beneficial for performance improvement on the OTB99\_Lang \cite{otb99} and MGIT \cite{mgit} benchmarks. For OTB99\_Lang \cite{otb99}, inaccurate natural language descriptions in official annotations create interference for tracking, while concise text provides further assurance for BBox that already expresses temporal and spatial relationships well, reducing interference. For MGIT \cite{mgit}, the algorithm is unable to understand complex temporal relationships and can only extract semantic information from concise text. Official text annotations of LaSOT \cite{lasot} lie between the two and are most conducive to the current algorithm, resulting in the best performance.

(2) \textbf{For short-term tracking task, dense concise text will bring greater gains. While dense detailed text is more suitable for the other two tasks.} Looking at the results of testing after retraining with different texts, dense concise text has the greatest impact on OTB99\_Lang \cite{otb99}. We think this is because the text provides precise object descriptions, further compensating for the shortcomings of BBox. The algorithm can further improve its performance on MGIT \cite{mgit} by learning from dense detailed text, because they can provide high-level semantic information that BBox cannot exhibit, such as temporal and spatial relationships. By text updating that best suits the memory system of algorithm, we provide the algorithm with precise and timely high-level semantic information, which is more helpful for understanding long video. 

(3) \textbf{The text processing method and multi-modal alignment ability need to be adjusted and improved.} The current algorithm cannot fully understand and learn complex temporal and spatial relationships. When the text processing and multi-modal alignment abilities of algorithm are adjusted and improved, text with more information will show even greater potential.
\begin{figure}[ht!]
  \centering
  \includegraphics[width=1\linewidth]{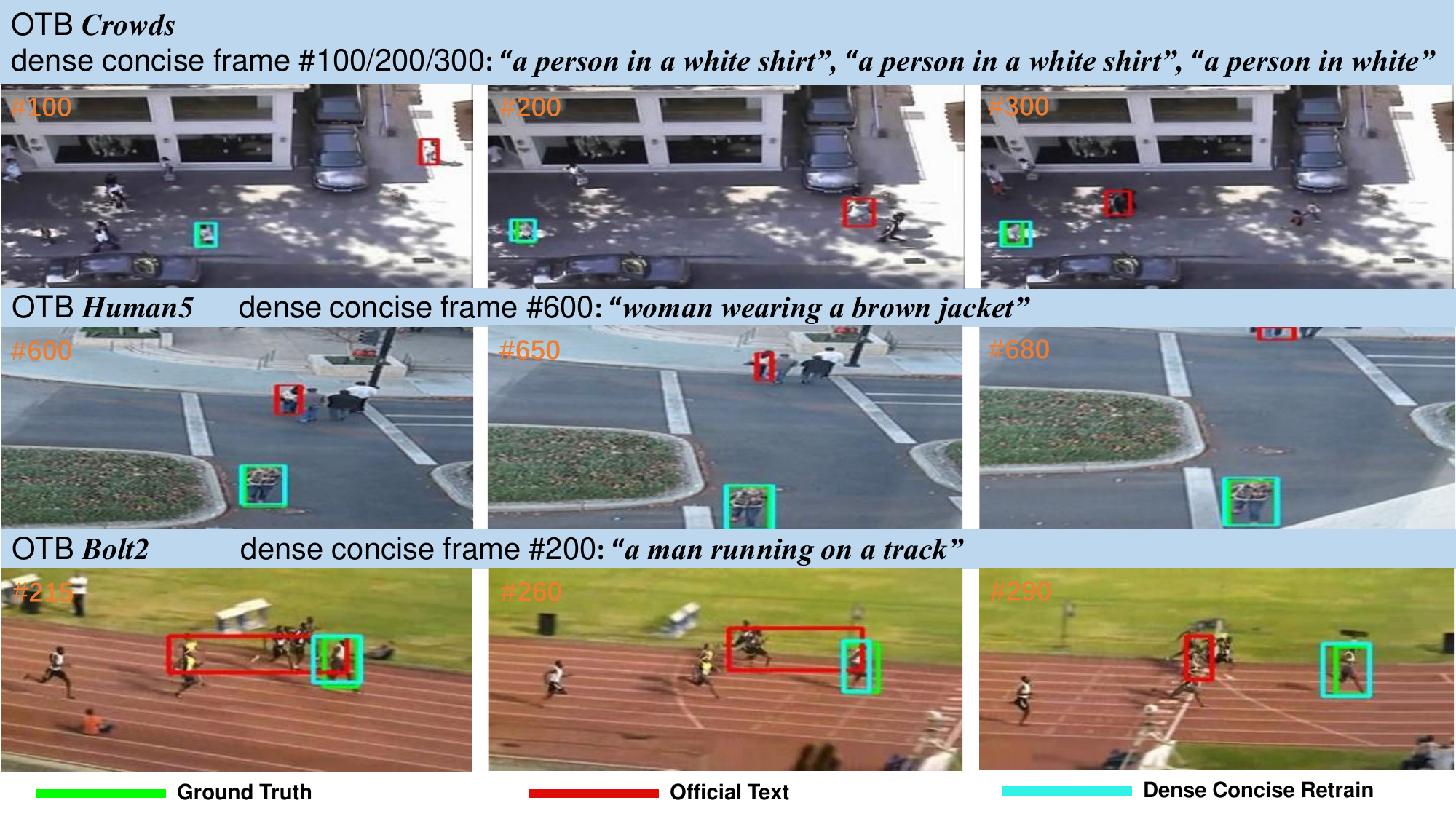}

  \caption{Visualization of tracking results on dense concise text annotations retrained algorithm.}
  \label{visualization}
\end{figure}
\subsection{Visualization}
As shown in Fig.~\ref{visualization}, we visualize the tracking results of the retrained model with official and dense concise text on three challenging sequences from OTB99\_Lang \cite{otb99}. In these sequences, the official text annotations can only cover a short time for the changes in the object. The scenes contain distractors, and the appearance of the object undergoes significant changes. The retrained model exhibits greater robustness with dense concise text compared to the official one. This validates that our generated text helps tracker to address these challenges. 




\section{Conclusions}
Object tracking is the basis for advanced tasks such as video understanding, and VLT may offer a potential path for enhancing tracking capabilities. In this paper, we propose DTLLM-VLT, a unified prompt framework, and generate diverse multi-granularity text descriptions. We analyze the results under different natural language descriptions for three representative benchmarks, aiming to provide new insights for the evaluation of different tracking tasks.

In our perspective, enhancing algorithm performance requires a comprehensive understanding of the properties of the datasets. We explore how leveraging the generative capabilities of LLM can help us improve VLT datasets and provide a new analytical approach from a multi-modal perspective for the field of video understanding. We hope this work can be expanded to incorporate more datasets, thereby enhancing support for vision datasets understanding research.

\section{Acknowledgement}
This work is jointly supported by the National Science and Technology Major Project (No.2022ZD0116403); the National Natural Science Foundation of China (No.62176255); the Strategic Priority Research Program of Chinese Academy of Sciences (No.XDA27000000).
{
    \small
    \bibliographystyle{ieeenat_fullname}
    \bibliography{main}

\begin{thebibliography}{10}
\providecommand{\url}[1]{#1}
\csname url@samestyle\endcsname
\providecommand{\newblock}{\relax}
\providecommand{\bibinfo}[2]{#2}
\providecommand{\BIBentrySTDinterwordspacing}{\spaceskip=0pt\relax}
\providecommand{\BIBentryALTinterwordstretchfactor}{4}
\providecommand{\BIBentryALTinterwordspacing}{\spaceskip=\fontdimen2\font plus
\BIBentryALTinterwordstretchfactor\fontdimen3\font minus \fontdimen4\font\relax}
\providecommand{\BIBforeignlanguage}[2]{{%
\expandafter\ifx\csname l@#1\endcsname\relax
\typeout{** WARNING: IEEEtran.bst: No hyphenation pattern has been}%
\typeout{** loaded for the language `#1'. Using the pattern for}%
\typeout{** the default language instead.}%
\else
\language=\csname l@#1\endcsname
\fi
#2}}
\providecommand{\BIBdecl}{\relax}
\BIBdecl

\bibitem{uav16}
M.~Mueller, N.~Smith, and B.~Ghanem, ``A benchmark and simulator for uav tracking,'' in \emph{Computer Vision--ECCV 2016: 14th European Conference, Amsterdam, The Netherlands, October 11--14, 2016, Proceedings, Part I 14}.\hskip 1em plus 0.5em minus 0.4em\relax Springer, 2016, pp. 445--461.

\bibitem{vot13}
R.~Bibliographie~Goecke, R.~Stolkin, S.~Lim, S.~Maher, S.~Poullot, S.~Wong, S.~Satoh, W.~Chen, W.~Hu, X.~Zhang \emph{et~al.}, ``The visual object tracking vot2013 challenge results,'' in \emph{Computer Vision Workshops (ICCVW), 2013 IEEE International Conference on}, 2013, pp. 98--111.

\bibitem{vot15}
M.~Kristan, J.~Matas, A.~Leonardis, M.~Felsberg, L.~Cehovin, G.~Fernandez, T.~Vojir, G.~Hager, G.~Nebehay, and R.~Pflugfelder, ``The visual object tracking vot2015 challenge results,'' in \emph{Proceedings of the IEEE international conference on computer vision workshops}, 2015, pp. 1--23.

\bibitem{vot18}
M.~Kristan, A.~Leonardis, J.~Matas, M.~Felsberg, R.~Pflugfelder, L.~ˇCehovin~Zajc, T.~Vojir, G.~Bhat, A.~Lukezic, A.~Eldesokey \emph{et~al.}, ``The sixth visual object tracking vot2018 challenge results,'' in \emph{Proceedings of the European conference on computer vision (ECCV) workshops}, 2018, pp. 0--0.

\bibitem{vot19}
M.~Kristan, J.~Matas, A.~Leonardis, M.~Felsberg, R.~Pflugfelder, J.-K. Kamarainen, L.~ˇCehovin~Zajc, O.~Drbohlav, A.~Lukezic, A.~Berg \emph{et~al.}, ``The seventh visual object tracking vot2019 challenge results,'' in \emph{Proceedings of the IEEE/CVF international conference on computer vision workshops}, 2019, pp. 0--0.

\bibitem{trackingnet}
M.~Muller, A.~Bibi, S.~Giancola, S.~Alsubaihi, and B.~Ghanem, ``Trackingnet: A large-scale dataset and benchmark for object tracking in the wild,'' in \emph{Proceedings of the European conference on computer vision (ECCV)}, 2018, pp. 300--317.

\bibitem{evaluate}
M.~Kristan, J.~Matas, A.~Leonardis, T.~Voj{\'\i}{\v{r}}, R.~Pflugfelder, G.~Fernandez, G.~Nebehay, F.~Porikli, and L.~{\v{C}}ehovin, ``A novel performance evaluation methodology for single-target trackers,'' \emph{IEEE transactions on pattern analysis and machine intelligence}, vol.~38, no.~11, pp. 2137--2155, 2016.

\bibitem{mgit}
S.~Hu, D.~Zhang, M.~Wu, X.~Feng, X.~Li, X.~Zhao, and K.~Huang, ``A multi-modal global instance tracking benchmark (mgit): Better locating target in complex spatio-temporal and causal relationship,'' in \emph{Advances in Neural Information Processing Systems}, vol.~36, 2023, pp. 25\,007--25\,030.

\bibitem{sotverse}
S.~Hu, X.~Zhao, and K.~Huang, ``Sotverse: A user-defined task space of single object tracking,'' \emph{International Journal of Computer Vision}, vol. 132, p. 872–930, 2024.

\bibitem{lasot}
H.~Fan, L.~Lin, F.~Yang, P.~Chu, G.~Deng, S.~Yu, H.~Bai, Y.~Xu, C.~Liao, and H.~Ling, ``Lasot: A high-quality benchmark for large-scale single object tracking,'' in \emph{Proceedings of the IEEE/CVF Conference on Computer Vision and Pattern Recognition}, 2019, pp. 5369--5378.

\bibitem{lasot21}
H.~Fan, H.~Bai, L.~Lin, F.~Yang, P.~Chu, G.~Deng, S.~Yu, Harshit, M.~Huang, J.~Liu \emph{et~al.}, ``Lasot: A high-quality large-scale single object tracking benchmark,'' \emph{International Journal of Computer Vision}, vol. 129, pp. 439--461, 2021.

\bibitem{tnl2k}
X.~Wang, X.~Shu, Z.~Zhang, B.~Jiang, Y.~Wang, Y.~Tian, and F.~Wu, ``Towards more flexible and accurate object tracking with natural language: Algorithms and benchmark,'' in \emph{Proceedings of the IEEE/CVF Conference on Computer Vision and Pattern Recognition}, 2021, pp. 13\,763--13\,773.

\bibitem{otb99}
Z.~Li, R.~Tao, E.~Gavves, C.~G. Snoek, and A.~W. Smeulders, ``Tracking by natural language specification,'' in \emph{Proceedings of the IEEE/CVF Conference on Computer Vision and Pattern Recognition}, 2017, pp. 6495--6503.

\bibitem{otb50}
Y.~Wu, J.~Lim, and M.-H. Yang, ``Online object tracking: A benchmark,'' in \emph{Proceedings of the IEEE/CVF Conference on Computer Vision and Pattern Recognition}, 2013, pp. 2411--2418.

\bibitem{otb100}
------, ``Object tracking benchmark,'' \emph{IEEE Transactions on Pattern Analysis and Machine Intelligence}, vol.~37, no.~09, pp. 1834--1848, 2015.

\bibitem{got10k}
L.~Huang, X.~Zhao, and K.~Huang, ``Got-10k: A large high-diversity benchmark for generic object tracking in the wild,'' \emph{IEEE Transactions on Pattern Analysis and Machine Intelligence}, vol.~43, no.~5, pp. 1562--1577, 2021.

\bibitem{oxuva}
J.~Valmadre, L.~Bertinetto, J.~F. Henriques, R.~Tao, A.~Vedaldi, A.~W. Smeulders, P.~H. Torr, and E.~Gavves, ``Long-term tracking in the wild: A benchmark,'' in \emph{Proceedings of the European conference on computer vision (ECCV)}, 2018, pp. 670--685.

\bibitem{git}
S.~Hu, X.~Zhao, L.~Huang, and K.~Huang, ``Global instance tracking: Locating target more like humans,'' \emph{IEEE Transactions on Pattern Analysis and Machine Intelligence}, vol.~45, no.~1, pp. 576--592, 2023.

\bibitem{biodrone}
X.~Zhao, S.~Hu, Y.~Wang, J.~Zhang, Y.~Hu, R.~Liu, H.~Ling, Y.~Li, R.~Li, K.~Liu \emph{et~al.}, ``Biodrone: A bionic drone-based single object tracking benchmark for robust vision,'' \emph{International Journal of Computer Vision}, pp. 1--26, 2023.

\bibitem{jointnlt}
L.~Zhou, Z.~Zhou, K.~Mao, and Z.~He, ``Joint visual grounding and tracking with natural language specification,'' in \emph{Proceedings of the IEEE/CVF Conference on Computer Vision and Pattern Recognition}, 2023, pp. 23\,151--23\,160.

\bibitem{snlt}
Q.~Feng, V.~Ablavsky, Q.~Bai, and S.~Sclaroff, ``Siamese natural language tracker: Tracking by natural language descriptions with siamese trackers,'' in \emph{Proceedings of the IEEE/CVF Conference on Computer Vision and Pattern Recognition}, 2021, pp. 5847--5856.

\bibitem{gti}
Z.~Yang, T.~Kumar, T.~Chen, J.~Su, and J.~Luo, ``Grounding-tracking-integration,'' \emph{IEEE Transactions on Circuits and Systems for Video Technology}, vol.~31, no.~9, pp. 3433--3443, 2021.

\bibitem{mmtrack}
Y.~Zheng, B.~Zhong, Q.~Liang, G.~Li, R.~Ji, and X.~Li, ``Towards unified token learning for vision-language tracking,'' \emph{IEEE Transactions on Circuits and Systems for Video Technology}, 2023.

\bibitem{vlt}
M.~Guo, Z.~Zhang, H.~Fan, and L.~Jing, ``Divert more attention to vision-language tracking,'' in \emph{Proceedings of the Advances in Neural Information Processing Systems}, S.~Koyejo, S.~Mohamed, A.~Agarwal, D.~Belgrave, K.~Cho, and A.~Oh, Eds., vol.~35, 2022, pp. 4446--4460.

\bibitem{transnlt}
R.~Wang, Z.~Tang, Q.~Zhou, X.~Liu, T.~Hui, Q.~Tan, and S.~Liu, ``Unified transformer with isomorphic branches for natural language tracking,'' \emph{IEEE Transactions on Circuits and Systems for Video Technology}, 2023.

\bibitem{transvlt}
H.~Zhao, X.~Wang, D.~Wang, H.~Lu, and X.~Ruan, ``Transformer vision-language tracking via proxy token guided cross-modal fusion,'' \emph{Pattern Recognition Letters}, vol. 168, pp. 10--16, 2023.

\bibitem{ctrtnl}
Y.~Li, J.~Yu, Z.~Cai, and Y.~Pan, ``Cross-modal target retrieval for tracking by natural language,'' in \emph{Proceedings of the IEEE/CVF Conference on Computer Vision and Pattern Recognition}, 2022, pp. 4931--4940.

\bibitem{dat}
X.~Wang, C.~Li, R.~Yang, T.~Zhang, J.~Tang, and B.~Luo, ``Describe and attend to track: Learning natural language guided structural representation and visual attention for object tracking,'' \emph{arXiv preprint arXiv:1811.10014}, 2018.

\bibitem{arxiv19}
Q.~Feng, V.~Ablavsky, Q.~Bai, and S.~Sclaroff, ``Robust visual object tracking with natural language region proposal network,'' \emph{arXiv preprint arXiv:1912.02048}, vol.~1, no.~7, p.~8, 2019.

\bibitem{rttnld}
Q.~Feng, V.~Ablavsky, Q.~Bai, G.~Li, and S.~Sclaroff, ``Real-time visual object tracking with natural language description,'' in \emph{Proceedings of the IEEE/CVF Winter Conference on Applications of Computer Vision}, 2020, pp. 700--709.

\bibitem{sam}
A.~Kirillov, E.~Mintun, N.~Ravi, H.~Mao, C.~Rolland, L.~Gustafson, T.~Xiao, S.~Whitehead, A.~C. Berg, W.-Y. Lo \emph{et~al.}, ``Segment anything,'' in \emph{Proceedings of the IEEE/CVF International Conference on Computer Vision}, 2023, pp. 4015--4026.

\bibitem{osprey}
Y.~Yuan, W.~Li, J.~Liu, D.~Tang, X.~Luo, C.~Qin, L.~Zhang, and J.~Zhu, ``Osprey: Pixel understanding with visual instruction tuning,'' \emph{arXiv preprint arXiv:2312.10032}, 2023.

\bibitem{llama}
H.~Touvron, T.~Lavril, G.~Izacard, X.~Martinet, M.-A. Lachaux, T.~Lacroix, B.~Rozi{\`e}re, N.~Goyal, E.~Hambro, F.~Azhar \emph{et~al.}, ``Llama: Open and efficient foundation language models,'' \emph{arXiv preprint arXiv:2302.13971}, 2023.

\bibitem{vicuna}
W.-L. Chiang, Z.~Li, Z.~Lin, Y.~Sheng, Z.~Wu, H.~Zhang, L.~Zheng, S.~Zhuang, Y.~Zhuang, J.~E. Gonzalez \emph{et~al.}, ``Vicuna: An open-source chatbot impressing gpt-4 with 90\%* chatgpt quality,'' \emph{See https://vicuna. lmsys. org (accessed 14 April 2023)}, vol.~2, no.~3, p.~6, 2023.

\bibitem{humanmemory}
G.~A. Radvansky, \emph{Human memory}.\hskip 1em plus 0.5em minus 0.4em\relax Routledge, 2021.

\bibitem{memory}
R.~D. Strous, N.~Cowan, W.~Ritter, and D.~C. Javitt, ``Auditory sensory ("echoic") memory dysfunction in schizophrenia.'' \emph{The American journal of psychiatry}, vol. 152, no.~10, pp. 1517--1519, 1995.

\bibitem{human}
R.~C. Atkinson and R.~M. Shiffrin, ``Human memory: A proposed system and its control processes,'' in \emph{Psychology of learning and motivation}.\hskip 1em plus 0.5em minus 0.4em\relax Elsevier, 1968, vol.~2, pp. 89--195.

\bibitem{siamrcnn}
P.~Voigtlaender, J.~Luiten, P.~H. Torr, and B.~Leibe, ``Siam r-cnn: Visual tracking by re-detection,'' in \emph{Proceedings of the IEEE/CVF conference on computer vision and pattern recognition}, 2020, pp. 6578--6588.

\bibitem{siamcar}
D.~Guo, J.~Wang, Y.~Cui, Z.~Wang, and S.~Chen, ``Siamcar: Siamese fully convolutional classification and regression for visual tracking,'' in \emph{Proceedings of the IEEE/CVF conference on computer vision and pattern recognition}, 2020, pp. 6269--6277.

\bibitem{prvt}
M.~Danelljan, L.~V. Gool, and R.~Timofte, ``Probabilistic regression for visual tracking,'' in \emph{Proceedings of the IEEE/CVF conference on computer vision and pattern recognition}, 2020, pp. 7183--7192.

\bibitem{keeptrack}
C.~Mayer, M.~Danelljan, D.~P. Paudel, and L.~Van~Gool, ``Learning target candidate association to keep track of what not to track,'' in \emph{Proceedings of the IEEE/CVF international conference on computer vision}, 2021, pp. 13\,444--13\,454.

\bibitem{mixformer}
Y.~Cui, C.~Jiang, L.~Wang, and G.~Wu, ``Mixformer: End-to-end tracking with iterative mixed attention,'' in \emph{Proceedings of the IEEE/CVF conference on computer vision and pattern recognition}, 2022, pp. 13\,608--13\,618.

\bibitem{siamrpn}
B.~Li, J.~Yan, W.~Wu, Z.~Zhu, and X.~Hu, ``High performance visual tracking with siamese region proposal network,'' in \emph{Proceedings of the IEEE conference on computer vision and pattern recognition}, 2018, pp. 8971--8980.

\end{thebibliography}
}


\end{document}